%% file: main.tex
\documentclass{article} 
\usepackage{iclr2023_conference_tinypaper,times}

\input{math_commands.tex}

\usepackage{hyperref}
\usepackage{url}
\usepackage{booktabs}
\usepackage{multirow}
\usepackage{graphicx}

\title{Explorations in Texture Learning}

\author{Blaine Hoak \\
University of Wisconsin-Madison\\
\texttt{bhoak@cs.wisc.edu} \\
\And{}Patrick McDaniel \\
University of Wisconsin-Madison\\
\texttt{mcdaniel@cs.wisc.edu} \\
}

\iclrfinalcopy 
\begin{document}

\maketitle
\begin{abstract}
In this work, we investigate \textit{texture learning}: the identification of textures learned by object classification models, and the extent to which they rely on these textures. We build texture-object associations that uncover new insights about the relationships between texture and object classes in CNNs and find three classes of results: associations that are strong and expected, strong and not expected, and expected but not present. Our analysis demonstrates that investigations in texture learning enable new methods for interpretability and have the potential to uncover unexpected biases. Code is available at \url{https://github.com/blainehoak/texture-learning}.
\end{abstract}

\section{Introduction and Background}

Convolutional Neural Networks (CNNs) have been shown to be more biased towards texture (repeated patterns), rather than shape like human vision is \citet{geirhos_imagenet-trained_2019}. Functionally, this suggests that models learn to associate object classes with textures that are present in the image, rather than shapes. This adherence to texture bias not only highlights discrepancies between human and machine vision, but may also impact model robustness and generalization \citet{geirhos_imagenet-trained_2019}. 

While prior works have focused on measuring, mitigating, and explaining texture bias in CNNs \citet{geirhos_imagenet-trained_2019, geirhos_partial_2021, hermann_origins_2020, gatys_texture_2015}, in this work we leverage the existence of texture bias to uncover what kinds of textures are learned. Specifically, we investigate \textit{texture learning}: the identification of textures learned by models during training and the extent to which these textures are associated with objects. We do this by building a mapping of texture-object associations, which allow us to understand (a) what kind of textures models may be biased toward and (b) when this has the potential to be problematic.

\section{Building Texture-Object Associations}
To uncover the textures that are learned by models, we build texture-object associations. Specifically, we input texture-only images into an ImageNet trained model and measure the degree to which certain textures are classified as specific objects. Importantly, and contrasting with prior work, the textures we explore are representative of texture classes that go beyond the typical textures that may be easily associated with ImageNet objects (e.g., elephant skin texture is easily associated with elephant objects, but bumpy textures do not readily map to one object class). This is to ensure that we remain free of assumptions about what textures ``should'' be associated with certain objects, and that we are capturing texture learning phenomena that may not be as expected.

To this end, we use the Describable Textures Dataset (DTD)~\cite{cimpoi_describing_2014} as our texture dataset. The DTD consists of 5640 images, each of which is labeled with one of 47 texture classes (e.g., bubbly, scaly, polka-dotted). We use a pretrained ResNet50 model (see Appendix~\ref{appendix_details} for details) to classify each of these texture images as belonging to one of the 1000 ImageNet classes (objects). \textbf{Notably, our experiments use a model trained on one dataset (ImageNet) yet are evaluated with an entirely different phenomenon (i.e., textures from DTD).}

With these texture classifications, we measure the effect size for each texture-object pairing (47 texture classes $\times$ 1000 object classes) by taking the ratio of samples belonging to the texture class that were classified as the corresponding ImageNet class. In other words, the effect size for texture class $A$ and object class $B$ represents how many samples belonging to texture class $A$ were predicted to be object class $B$. Thus, higher effect sizes correspond to stronger texture-object associations. For each of the 47 texture classes, the top 3 objects classes with the highest effect size are reported. \autoref{class_mappings} of Appendix~\ref{appendix} contains the texture-object associations for all 47 texture classes (3 object classes per texture class), sorted highest to lowest by the first effect size column. For brevity, \autoref{class_mappings_short} displays the first 10 rows of this table, corresponding to the top 10 textures with the strongest associations. 

\begin{table}[tp]
    \centering
    \resizebox{.92\textwidth}{!}{\begin{minipage}{\textwidth}
        \begin{center}
            \input{texture_imagenet_top3_normalize_short}
        \end{center}
    \end{minipage}}
    \caption{First 10 rows of the texture-object associations with the top 3 most predicted objects (and their effect size) for each texture.}\label{class_mappings_short}
  \end{table}

The texture-object associations yield multiple interesting results, which we divide into three types based on (a) how expected the relationship between texture and object is (i.e., if humans would naturally associate the texture with the object) and (b) the strength of the association that emerged in our results. Below we provide examples and descriptions of each type. See corresponding sections of Appendix \ref{appendix} for images of each example. 

\noindent\textbf{Expected \& Strongly Present.} The \textit{honeycombed} textures (which consist of repeated hexagonal patterns in objects ranging from bathroom tiles to bee honeycombs) were classified as the \textit{honeycomb} object 73.1\% of the time. This is a strong association, and is not necessarily surprising, as honeycomb objects are largely composed of honeycombed textures. Despite the ``expectedness'' of the association, these results are still interesting for two reasons. First, this demonstrates that models are able to generalize well on textures alone for these object classes, even for examples that are in entirely different datasets. Second, given that the DTD honeycombed texture images consisted of a variety of objects beyond honeycombs, the strength of this association suggests that the ImageNet model is predominantly relying on texture to predict the classes of these categories of images, rather than color or shape. See Appendix \ref{exp_pres} for supporting images.

\noindent\textbf{Not Expected \& Strongly Present.} The \textit{polka-dotted} and \textit{dotted} texture classes were most often mapped to the \textit{bib} object (24.8 \% and 24.7\% of the time, respectively). While there is not an obvious object class that the polka-dotted or dotted textures would naturally be associated with, the strength of this association suggests that the model has indeed learned to associate these textures with the bib object. This could suggest a bias in the training data that was learned by the model: a large number of training examples for the bib object may contain polka-dotted or dotted textures. In subsequent investigations of the ImageNet training data, we found this to be true; a glance through some of the \textit{bib} images recovered multiple examples of bibs with polka-dots (shown in \autoref{fig:dotted_bibs}) See \ref{noexp_pres}.

\noindent\textbf{Expected \& Not Present.} The \textit{scaly} texture images, while consisting of images appearing to be fish and reptile scales, were not associated with any fish or reptile objects, but rather with the honeycomb object (13.5\% of the time as shown in \autoref{class_mappings}). These types of results highlight object classes that may not have learned generalizable textures. This could be due to the fact that the textures in the training examples of these objects were not diverse enough, or that these object classes learned to build stronger associations to shapes or colors, rather than textures. See \ref{exp_nopres} for supporting images.

\section{Conclusions}
This methodology and subsequent findings can be used to uncover learned (and potentially unexpected) associations. This not only enables greater model interpretability, but can also highlight and identify specific unwanted biases in models. 

\subsubsection*{URM Statement}
The authors acknowledge that at least one key author of this work meets the URM criteria of ICLR 2024 Tiny Papers Track.

\section*{Acknowledgements}
This material is based upon work supported by, or in part by, the National Science Foundation under Grant No. CNS1946022 and Grant No. CNS2343611. Any opinions, findings, and conclusions or recommendations expressed in this publication are those of the author(s) and do not necessarily reflect the views of the National Science Foundation, or the U.S. Government. The U.S. Government is authorized to reproduce and distribute reprints for government purposes notwithstanding any copyright notation hereon.

\bibliography{references}
\bibliographystyle{iclr2023_conference_tinypaper}

\appendix
\section{Appendix}\label{appendix}
\subsection{Experimental Details}\label{appendix_details}
The pretrained ResNet50 used in our experiments was obtained from torchvision~\cite{marcel_torchvision_2010} with the default model weights. The model was trained on ImageNet~\cite{russakovsky_imagenet_2015}. The model was evaluated on the DTD dataset using the following data preprocessing steps: (1) resize the image to 256$\times$256, (2) center crop the image to 224$\times$224, (3) normalize the image using the mean and standard deviation of the ImageNet training dataset. All experiments were run on a single NVIDIA 2080Ti GPU. Complete code to replicate experiments can be found at \url{https://github.com/blainehoak/texture-learning}.

\subsection{Image Examples}
\subsubsection{Expected \& Strongly Present}\label{exp_pres}
Images associated with the honeycombed class of the Describable Textures Dataset can be browsed at \url{https://www.robots.ox.ac.uk/~vgg/data/dtd/view.html?categ=honeycombed}. 

\subsubsection{Not Expected \& Strongly Present}\label{noexp_pres}
Images associated with the dotted and polka dotted classes of the Describable Textures Dataset can be browsed at \url{https://www.robots.ox.ac.uk/~vgg/data/dtd/view.html?categ=dotted} and \url{https://www.robots.ox.ac.uk/~vgg/data/dtd/view.html?categ=polka_dotted}, respectively. 

\begin{figure}[h]
\begin{center}
\includegraphics[width=0.9\textwidth]{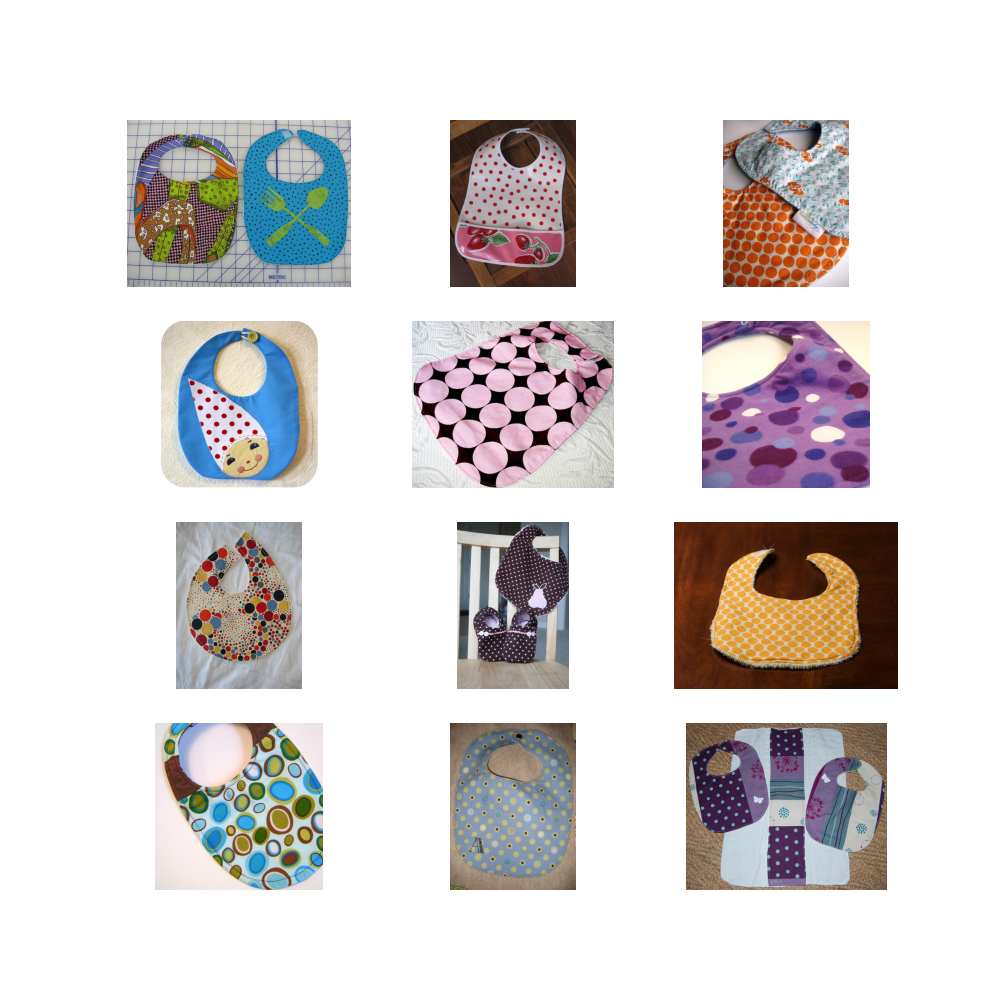}
\end{center}
\caption{Examples of bibs with polka-dots in the ImageNet training data.}
\label{fig:dotted_bibs}
\end{figure}

Upon further inspection of a portion of the ImageNet training data, we were able to easily find multiple examples where images in the \textit{bib} class contained dots or polka-dots. A few examples are shown in \autoref{fig:dotted_bibs}. This supports our hypothesis that the model may have learned a bias for polka-dots in the bib class, leading to strong object-texture associations in our results. This finding demonstrates that texture learning analysis may be a fruitful direction for uncovering biases in models.

\subsubsection{Expected \& Not Present}\label{exp_nopres}
Images associated with the scaly class of the Describable Textures Dataset can be browsed at \url{https://www.robots.ox.ac.uk/~vgg/data/dtd/view.html?categ=scaly}. 

\subsection{Extended Results}
\autoref{class_mappings} shows the full table of texture-object associations for all 47 texture classes on ResNet50. Results on Resnet152 can be found in \autoref{class_mappings_resnet152}
\begin{table}[htbp]
    \centering
    \resizebox{.92\textwidth}{!}{\begin{minipage}{\textwidth}
        \begin{center}
            \input{texture_imagenet_top3_normalize}
        \end{center}
    \end{minipage}}
    \caption{Texture-object associations with the top 3 most predicted objects (and their effect size) for each texture.}\label{class_mappings}
  \end{table}
\begin{table}[htbp]
    \centering
    \resizebox{.92\textwidth}{!}{\begin{minipage}{\textwidth}
        \begin{center}
            \input{texture_imagenet_top3_resnet152}
        \end{center}
    \end{minipage}}
    \caption{Texture-object associations with the top 3 most predicted objects (and their effect size) for each texture on ResNet152.}\label{class_mappings_resnet152}
  \end{table}

\end{document}

%% file: math_commands.tex
\usepackage{amsmath,amsfonts,bm}

\def\eqref#1{equation~\ref{#1}}

\def\1{\bm{1}}

\DeclareMathAlphabet{\mathsfit}{\encodingdefault}{\sfdefault}{m}{sl}
\SetMathAlphabet{\mathsfit}{bold}{\encodingdefault}{\sfdefault}{bx}{n}



%% file: texture_imagenet_top3_normalize_short.tex
\begin{tabular}{l|lrlrlr}
    \toprule
    Texture class & Object class & Effect & Object class & Effect & Object class & Effect \\
    \midrule
    honeycombed & honeycomb & 0.731 & chain\_mail & 0.071 & velvet & 0.027 \\
    cobwebbed & spider\_web & 0.655 & poncho & 0.046 & radio\_telescope & 0.046 \\
    waffled & waffle\_iron & 0.427 & honeycomb & 0.117 & pretzel & 0.075 \\
    striped & zebra & 0.381 & tiger & 0.169 & velvet & 0.093 \\
    knitted & dishrag & 0.331 & wool & 0.239 & cardigan & 0.188 \\
    stratified & cliff & 0.305 & velvet & 0.140 & stone\_wall & 0.125 \\
    spiralled & coil & 0.296 & maze & 0.061 & chambered\_nautilus & 0.043 \\
    bubbly & bubble & 0.286 & beer\_glass & 0.104 & Petri\_dish & 0.077 \\
    dotted & bib & 0.248 & shower\_curtain & 0.148 & wallet & 0.097 \\
    polka-dotted & bib & 0.247 & Windsor\_tie & 0.125 & wallet & 0.089 \\
    \bottomrule
    \end{tabular}

%% file: texture_imagenet_top3_normalize.tex
\begin{tabular}{l|l r l r l r}
    \toprule
    Texture class & Object class & Effect & Object class & Effect & Object class & Effect \\
    \midrule
    honeycombed & honeycomb & 0.731 & chain mail & 0.071 & velvet & 0.027 \\
    cobwebbed & spider web & 0.655 & poncho & 0.046 & radio telescope & 0.046 \\
    waffled & waffle iron & 0.427 & honeycomb & 0.117 & pretzel & 0.075 \\
    striped & zebra & 0.381 & tiger & 0.169 & velvet & 0.093 \\
    knitted & dishrag & 0.331 & wool & 0.239 & cardigan & 0.188 \\
    stratified & cliff & 0.305 & velvet & 0.140 & stone wall & 0.125 \\
    spiralled & coil & 0.296 & maze & 0.061 & chambered nautilus & 0.043 \\
    bubbly & bubble & 0.286 & beer glass & 0.104 & Petri dish & 0.077 \\
    dotted & bib & 0.248 & shower curtain & 0.148 & wallet & 0.097 \\
    polka-dotted & bib & 0.247 & Windsor tie & 0.125 & wallet & 0.089 \\
    paisley & velvet & 0.223 & wool & 0.112 & shower curtain & 0.103 \\
    wrinkled & velvet & 0.219 & quilt & 0.153 & wool & 0.051 \\
    frilly & head cabbage & 0.209 & hoopskirt & 0.105 & velvet & 0.069 \\
    grid & window screen & 0.199 & oscilloscope & 0.114 & shoji & 0.063 \\
    crystalline & plastic bag & 0.193 & head cabbage & 0.082 & honeycomb & 0.068 \\
    lacelike & handkerchief & 0.191 & velvet & 0.119 & stole & 0.108 \\
    perforated & strainer & 0.190 & space heater & 0.080 & honeycomb & 0.074 \\
    stained & velvet & 0.184 & volcano & 0.040 & potpie & 0.035 \\
    woven & hamper & 0.175 & velvet & 0.156 & dishrag & 0.100 \\
    blotchy & velvet & 0.164 & ant & 0.058 & fig & 0.032 \\
    gauzy & shower curtain & 0.158 & velvet & 0.079 & window shade & 0.068 \\
    cracked & stone wall & 0.158 & guillotine & 0.074 & spider web & 0.074 \\
    braided & knot & 0.155 & hamper & 0.125 & dishrag & 0.097 \\
    zigzagged & maze & 0.153 & envelope & 0.131 & quilt & 0.115 \\
    meshed & chainlink fence & 0.148 & honeycomb & 0.140 & window screen & 0.137 \\
    interlaced & maze & 0.148 & prayer rug & 0.092 & shield & 0.065 \\
    veined & leaf beetle & 0.143 & head cabbage & 0.095 & sulphur butterfly & 0.049 \\
    lined & shower curtain & 0.142 & web site & 0.094 & window shade & 0.073 \\
    banded & shower curtain & 0.142 & bib & 0.079 & Windsor tie & 0.079 \\
    marbled & velvet & 0.137 & cliff & 0.052 & spider web & 0.044 \\
    flecked & wool & 0.135 & velvet & 0.080 & cardigan & 0.069 \\
    scaly & honeycomb & 0.135 & tile roof & 0.071 & wool & 0.061 \\
    matted & wool & 0.132 & komondor & 0.070 & wig & 0.059 \\
    pleated & shower curtain & 0.129 & velvet & 0.118 & window shade & 0.102 \\
    crosshatched & window screen & 0.127 & velvet & 0.069 & handkerchief & 0.066 \\
    fibrous & hay & 0.126 & pot & 0.076 & matchstick & 0.050 \\
    swirly & fire screen & 0.116 & velvet & 0.103 & shower curtain & 0.084 \\
    grooved & radiator & 0.115 & velvet & 0.100 & doormat & 0.084 \\
    porous & French loaf & 0.115 & honeycomb & 0.049 & velvet & 0.044 \\
    chequered & wool & 0.114 & tray & 0.108 & crossword puzzle & 0.079 \\
    studded & strainer & 0.110 & Windsor tie & 0.105 & cuirass & 0.059 \\
    potholed & volcano & 0.108 & geyser & 0.090 & cliff dwelling & 0.063 \\
    freckled & lipstick & 0.104 & seat belt & 0.083 & Band Aid & 0.064 \\
    sprinkled & ice cream & 0.075 & dough & 0.070 & pretzel & 0.052 \\
    bumpy & custard apple & 0.073 & jackfruit & 0.049 & spaghetti squash & 0.047 \\
    pitted & pomegranate & 0.068 & doormat & 0.047 & switch & 0.042 \\
    smeared & mask & 0.057 & velvet & 0.054 & jellyfish & 0.041 \\
    \bottomrule
    \end{tabular}

%% file: texture_imagenet_top3_resnet152.tex
\begin{tabular}{l|l r l r l r}
    \toprule
    Texture class & Object class & Effect & Object class & Effect & Object class & Effect \\
    \midrule
    honeycombed & honeycomb & 0.753 & Christmas\_stocking & 0.026 & coil & 0.026 \\
    cobwebbed & spider\_web & 0.691 & barn\_spider & 0.074 & shower\_curtain & 0.025 \\
    waffled & waffle\_iron & 0.533 & honeycomb & 0.078 & tile\_roof & 0.044 \\
    spiralled & coil & 0.471 & maze & 0.058 & knot & 0.038 \\
    striped & zebra & 0.426 & tiger & 0.167 & velvet & 0.056 \\
    bubbly & bubble & 0.425 & beer\_glass & 0.085 & honeycomb & 0.057 \\
    knitted & dishrag & 0.417 & wool & 0.157 & cardigan & 0.148 \\
    dotted & bib & 0.415 & shower\_curtain & 0.113 & wallet & 0.066 \\
    polka-dotted & bib & 0.340 & pillow & 0.078 & shower\_curtain & 0.068 \\
    stratified & cliff & 0.309 & cliff\_dwelling & 0.082 & velvet & 0.073 \\
    wrinkled & velvet & 0.257 & quilt & 0.115 & packet & 0.044 \\
    zigzagged & pillow & 0.243 & maze & 0.122 & wool & 0.113 \\
    grid & window\_screen & 0.243 & manhole\_cover & 0.043 & oscilloscope & 0.043 \\
    paisley & velvet & 0.235 & shower\_curtain & 0.165 & pillow & 0.087 \\
    lacelike & handkerchief & 0.221 & quilt & 0.142 & shower\_curtain & 0.115 \\
    meshed & window\_screen & 0.198 & honeycomb & 0.153 & chainlink\_fence & 0.117 \\
    potholed & volcano & 0.195 & manhole\_cover & 0.161 & valley & 0.068 \\
    banded & shower\_curtain & 0.190 & web\_site & 0.155 & bib & 0.129 \\
    gauzy & shower\_curtain & 0.183 & velvet & 0.139 & window\_shade & 0.096 \\
    chequered & wool & 0.179 & shower\_curtain & 0.103 & wall\_clock & 0.085 \\
    perforated & window\_screen & 0.179 & strainer & 0.170 & honeycomb & 0.054 \\
    woven & hamper & 0.177 & dishrag & 0.097 & doormat & 0.080 \\
    frilly & head\_cabbage & 0.169 & gown & 0.102 & vase & 0.059 \\
    matted & wool & 0.169 & wig & 0.119 & komondor & 0.042 \\
    pleated & shower\_curtain & 0.157 & window\_shade & 0.139 & wool & 0.087 \\
    braided & knot & 0.155 & hamper & 0.121 & wool & 0.103 \\
    fibrous & hay & 0.139 & wool & 0.070 & pot & 0.052 \\
    veined & leaf\_beetle & 0.139 & head\_cabbage & 0.104 & buckeye & 0.087 \\
    lined & web\_site & 0.138 & wool & 0.095 & shower\_curtain & 0.095 \\
    blotchy & velvet & 0.137 & ant & 0.043 & switch & 0.043 \\
    crosshatched & window\_screen & 0.134 & wallet & 0.042 & handkerchief & 0.042 \\
    grooved & radiator & 0.127 & velvet & 0.059 & doormat & 0.059 \\
    freckled & lipstick & 0.120 & seat\_belt & 0.068 & cellular\_telephone & 0.051 \\
    marbled & velvet & 0.119 & cliff & 0.068 & spider\_web & 0.042 \\
    porous & French\_loaf & 0.118 & stone\_wall & 0.042 & manhole\_cover & 0.042 \\
    crystalline & plastic\_bag & 0.117 & head\_cabbage & 0.108 & shower\_cap & 0.058 \\
    studded & strainer & 0.111 & Windsor\_tie & 0.068 & purse & 0.060 \\
    flecked & cardigan & 0.103 & wool & 0.095 & velvet & 0.052 \\
    cracked & volcano & 0.103 & stone\_wall & 0.077 & tiger\_beetle & 0.077 \\
    sprinkled & ice\_cream & 0.102 & Petri\_dish & 0.042 & tray & 0.042 \\
    bumpy & custard\_apple & 0.101 & jackfruit & 0.059 & thimble & 0.034 \\
    scaly & honeycomb & 0.094 & wool & 0.068 & tile\_roof & 0.068 \\
    stained & velvet & 0.093 & paper\_towel & 0.059 & jean & 0.034 \\
    interlaced & maze & 0.093 & prayer\_rug & 0.085 & doormat & 0.059 \\
    swirly & shower\_curtain & 0.087 & fire\_screen & 0.061 & pillow & 0.061 \\
    smeared & paintbrush & 0.050 & mask & 0.042 & handkerchief & 0.042 \\
    pitted & switch & 0.050 & pomegranate & 0.042 & ant & 0.034 \\
    \bottomrule
    \end{tabular}